\pdfoutput=1
\documentclass[11pt]{article}

\usepackage[final]{acl}

\usepackage{times}
\usepackage{latexsym}

\usepackage[T1]{fontenc}

\usepackage[utf8]{inputenc}

\usepackage{microtype}

\usepackage{inconsolata}

\usepackage{graphicx}
\usepackage{algorithm}
\usepackage{algorithmic}
\usepackage{color}
\usepackage{bbding}
\usepackage{tikz}
\usepackage{lipsum}
\usepackage[most]{tcolorbox}
\usepackage{tikz}
\usepackage{comment}
\usepackage{booktabs}
\usepackage{multirow}
\usepackage{enumitem} 
\usepackage{float}
\usepackage{colortbl}
\usepackage{xcolor}
\usepackage{siunitx}
\title{P3: Prompts Promote Prompting}

\author{
  \textbf{Xinyu Zhang\textsuperscript{1,\dag}},
  \textbf{Yuanquan Hu\textsuperscript{2,\dag}},
  \textbf{Fangchao Liu\textsuperscript{2,\dag}},
  \textbf{Zhicheng Dou\textsuperscript{1,*}},
\\
  \textsuperscript{1}Renmin University of China,
  \textsuperscript{2}Huawei Poisson Lab
\\
 zhangxinyu1995@ruc.edu.cn, dou@ruc.edu.cn
}

\begin{document}
\maketitle

\def\thefootnote{\dag}\footnotetext{These authors contributed equally to this work.}\def\thefootnote{\arabic{footnote}}
\def\thefootnote{*}\footnotetext{Corresponding author.}\def\thefootnote{\arabic{footnote}}

\begin{abstract}
Current large language model (LLM) applications often employ multi-component prompts, comprising both system and user prompts, to guide model behaviors. 
While recent advancements have demonstrated the efficacy of automatically optimizing either the system or user prompt to boost performance, such unilateral approaches often yield suboptimal outcomes due to the interdependent nature of these components. 
In this work, we introduce P3, a novel self-improvement framework that concurrently optimizes both system and user prompts through an iterative process. The offline optimized prompts are further leveraged to promote online prompting by performing query-dependent prompt optimization. Extensive experiments on general tasks (e.g., Arena-hard and Alpaca-eval) and reasoning tasks (e.g., GSM8K and GPQA) demonstrate that P3 achieves superior performance in the realm of automatic prompt optimization.
Our results highlight the effectiveness of a holistic optimization strategy in enhancing LLM performance across diverse domains. 
\end{abstract}

\section{Introduction}
Large language models (LLMs) have achieved remarkable success across a wide range of tasks, becoming a cornerstone of modern AI applications. However, finetuning these models is often costly and data-dependent, driving interest in `external' tuning methods, e.g., prompt engineering and retrieval-augmented generation, which optimize outputs without retraining. 

Among these methods, prompt engineering stands out as a cost-effective and versatile approach, enabling significant performance gains with minimal adjustments. Yet, crafting effective prompts remains challenging, often requiring expertise and iterative manual tuning. Recent advances in automated prompt optimization (APO)~\citep{Dspy,EPO,Promptbreeder,APO,APE} address this by leveraging LLMs to refine prompts based on objective metrics or subjective feedbacks (LLM-as-judge). While these methods show solid results on several benchmarks, they lack real-time applicability, because their reliance on multiple model calls for each prompt refinement can introduce huge computational overhead.

\begin{figure}[t]
\centering
  \includegraphics[width=0.5\textwidth]{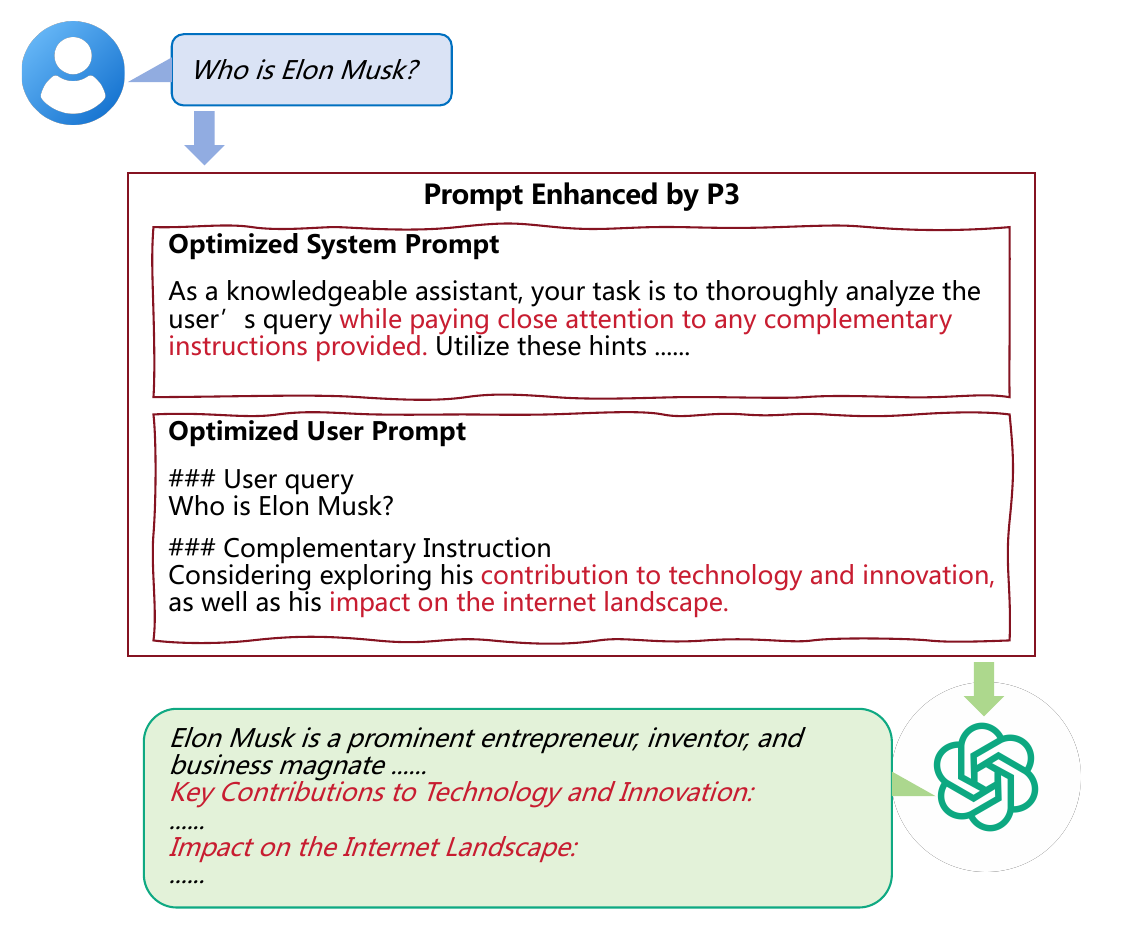}
  \caption {An illustration of P3's optimization results.}
  \vspace{-1em}
\label{fig:overview}
\end{figure}

To enable real-time prompt optimization, a new class of techniques for online prompt optimization has emerged, such as BPO~\citep{BPO} and PAS~\citep{PAS}. These methods can be divided into two stages. In the first stage, they optimize each user prompt and collect (prompt, optimized prompt) pairs as datasets. Specifically, BPO optimizes user prompts by rephrasing, and PAS adds a complementary instruction as an expansion to the original prompt. In the second stage, the collected dataset is used to finetune a smaller language model, which is capable of dynamically adjusting or extending the original prompt at test time. While promising, these approaches encounter three key limitations:

\textbf{Affinity Issue}: System prompts significantly influence the style and quality of LLM outputs. Existing methods primarily focus on optimizing user prompts, overlooking the potential of system prompts. To fully unlock the capabilities of prompt optimization, system prompts must adapt to the structural changes introduced by optimized user prompts, ensuring alignment and synergy between the two components.

\textbf{Diversity Issue}: There are numerous possible ways to rewrite or extend a prompt, making it difficult to ensure that any single adjustment is optimal. Current methods rely on a fixed rephrasing or extension strategy, which can introduce bias and lack exploration for global optimal solutions.

\textbf{Inference Cost and Efficiency Issue}: Existing online prompt optimization methods require finetuning smaller LLMs (e.g., 7B instruct models). This still necessitates substantial resources. And since these models operate serially alongside the primary LLM, they introduce additional inference latency and cost, further hindering real-time applications.

In response to these challenges, we propose Prompts Promote Prompting~(P3), which is a two-stage prompt optimization framework leveraging offline \textbf{prompts} to \textbf{promote} online \textbf{prompting}. Given that system prompts are more suited for offline optimization, while user prompts need to be dynamically adjusted based on specific queries, our method leverages both offline joint optimization for system prompts and real-time user prompt complement for more flexible and efficient application, as shown in Figure~\ref{fig:overview}.

For the \textbf{affinity issue}, we jointly optimize system prompts and user prompt complements in the offline stage. This involves iteratively refining the system prompt to ensure it effectively guides the model’s behavior and adapts to the paradigm of user prompt complementation, ensuring both system and user prompts work synergistically. For the \textbf{diversity issue}, we synthesize diverse user prompt complements in the offline optimization phase. This strategy ensures that user prompt complements cover diverse linguistic and contextual expressions, improving the flexibility and generalization of prompt optimization. Finally, for \textbf{inference efficiency}, we provide an alternative usage of the collected offline dataset in the online stage. Apart from finetuning a small LLM to perform online optimization, we provide an alternative usage of the dataset: retrieving samples from the dataset as demonstrations for in-context learning. This allows for prompt adjustments in real time while minimizing computational costs and inference time.

To validate the effectiveness of our method, we conducted comprehensive experiments across various general and reasoning tasks, including Arean-Hard~\cite{Arena-Hard}, Alpaca-Eval 2.0~\cite{alpaca_eval}, Alpaca-Eval (LC)~\cite{alpaca-eval-lc}, GSM8k~\cite{cobbe2021gsm8k} and GPQA~\cite{rein2024gpqa}. The evaluation is performed on multiple LLMs, including GPT-4o~\cite{gpt-4o}, GPT-4~\cite{openai2024gpt4}, GPT-4-turbo, GPT-3.5-turbo, Qwen-2~\cite{yang2024qwen2}, and LLaMA-3~\cite{meta2024llama3}. The results demonstrate that P3 consistently outperforms existing prompt optimization methods across all evaluated models. Additionally, the in-context learning version of P3 (P3-ICL) achieves a good balance between inference efficiency and performance gains. These findings highlight the robustness and versatility of P3 in enhancing LLM performance in real-time applications.

\section{Methodology}

\begin{figure*}[htb]
\vspace{-1em}
\centering
  \includegraphics[width=1.0\textwidth]{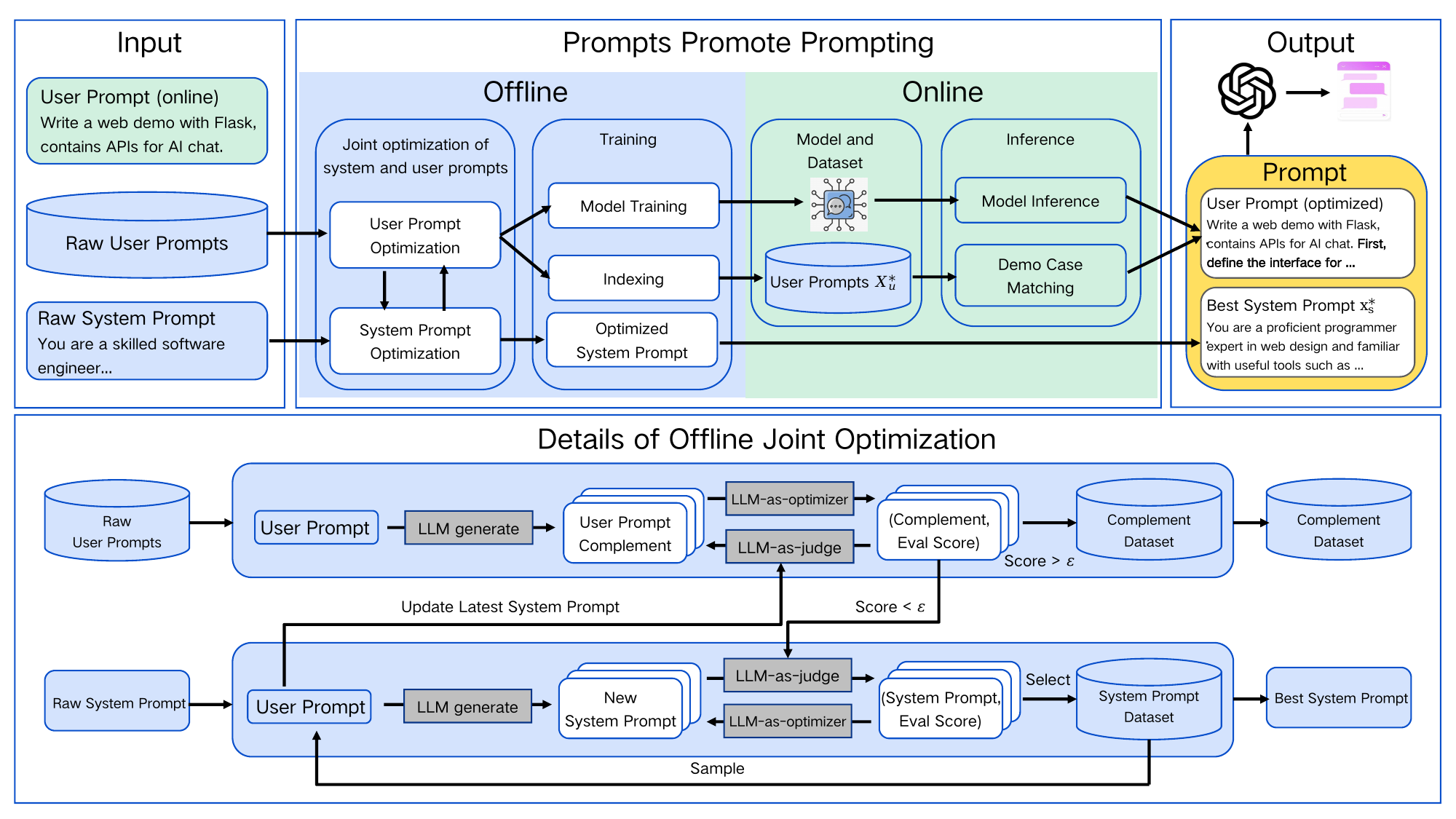}
  \caption {The overall framework of our methods. The complete process includes the offline joint optimization for user prompt and system prompt, which iteratively optimizes the system prompt and then generates the complementary user prompts. The generated user prompts are fed to train a smaller model for online prompt optimization or serve as a dataset for in-context learning. At test time, a user query is enhanced by a complement generated by the online optimization model or few-shot demonstrations retrieved from the offline dataset, guiding LLMs to better respond to the user query. The details of the offline joint optimization process are shown in the lower part of the figure.}
\label{fig:methods}
\end{figure*}

In this section, we introduce the background and overall framework of P3. We first give the task formulation for automatic prompt optimization in Section~\ref{sec:task}. Then we describe the algorithm details of P3 in section~\ref{sec:p3}.

\subsection{Task Definition}
\label{sec:task}
In this work, we address black-box prompt optimization~\citep{PAS,BPO}, formulated as:
\begin{equation}
x_{\text{opt}} = \mathcal{F}(x),
\end{equation}
where $x$ denotes an input prompt and $x_{\text{opt}}$ its optimized counterpart, enhancing LLM output quality. Our objective is to develop a black-box function $\mathcal{F}$ mapping $x$ to $x_{\text{opt}}$. Drawing inspiration from generative optimization approaches~\citep{yuksekgonul2024textgrad, opro,cheng2024trace}, we implement $\mathcal{F}$ using an LLM-driven system.

In current LLM applications, prompts are logically divided into two parts: the system prompt $x_s$, which defines the role, capability, or other global constraints for LLMs to follow, and the user prompt $x_u$, which gives the concrete requirements for the current conversation. However, previous methods mostly focus on either system prompt or user prompt optimization, while keeping the other part fixed. It may lead to sub-optimal optimization results because those two parts are intertwined in LLMs' decision process. Therefore, a holistic prompt optimization framework needs to consider the optimization of both system prompts and user prompts. To alleviate this affinity issue, we propose to jointly optimize the system and user prompts.


%

\subsection{Prompts Promote Prompting (P3)}
\label{sec:p3}
Figure~\ref{fig:methods} illustrates the complete process of P3. We employ different optimization strategies for system and user prompts, considering their different characteristics. System prompts, which define knowledge boundaries and establish behavioral guidelines for a given scenario, are optimized offline using a representative dataset for that scenario. This offline optimization allows for a fixed system prompt during online deployment, ensuring consistent behavior. User prompts, however, are inherently dynamic and context-dependent, necessitating adaptive online optimization. P3 amortizes the learning of the online user prompt optimizer during the offline stage, where user prompts are jointly optimized alongside system prompts. These optimized user prompts are then compiled into a dataset, which serves as a foundation for efficient and dynamic online adaptation. 


\subsubsection{User Prompt Optimization}
\label{sec:usr}
For user prompt optimization, we follow \citet{PAS} to generate a piece of complementary instruction that provides feasible direction and a short thought process for answering the user prompt.

In the offline stage, we iteratively optimize the complementary instruction for each user prompt. As shown in Figure~\ref{fig:methods}, for each user prompt, $k$ candidate instructions are generated by LLMs as the initial population. We concatenate these candidates with the user prompt and system prompt, and prompt LLMs to generate answers. The answers are evaluated using LLM-as-judge. The candidates and their scores serve as few-shot demonstrations for LLM-as-optimizer~\citep{opro} to generate refined instructions. This iterative process could continue for multiple rounds, with each round's optimized instructions selected as demonstrations for the next. This optimization procedure ensures both width (few-shot generation) and depth (multi-round iteration), enhancing exploration in the solution space, and thus addressing the aforementioned diversity issue. 

The highest-scoring complementary instruction is selected as the final label for the user prompt. If the final score is above a threshold $\epsilon$, we collect the (user prompt, complementary instruction) pair, appending it to the dataset $X^{*}_{u}$ used for finetuning the online optimization model. Otherwise, it is appended to a hard sample buffer $X_{u}$ for periodic system prompt optimization. 

\subsubsection{System Prompt Optimization}
\label{sec:sys}
For system prompt optimization, we leverage the hard samples identified during user prompt optimization. These samples, which are of lower quality and unsuitable for online model fine-tuning, are repurposed to enhance the resilience of system prompts and improve the overall data efficiency of our framework.

The optimization process for system prompts follows an LLM-as-optimizer approach, similar to that used for user prompts. We first initialize a buffer of system prompt $X_s$ based on the raw system prompt $x_s$ by prompting an LLM to refine it. Each time we sample a batch from the hard sample buffer and $k$ system prompts from $X_s$, and evaluate each system prompt over the batch. Then the system prompts and their scores (average score over the batch) serve as few-shot demonstrations to generate refined system prompts. Finally, the refined system prompts with the high scores are appended to $X_s$. And the best one is updated as the current system prompt for subsequent user prompt optimization, ensuring continuous improvement in system prompt effectiveness. The pseudocodes of the offline optimization procedure are provided in Appendix~\ref{appendix:algo}. And we provide meta prompts in Appendix~\ref{appendix:meta}.

\subsubsection{Online Optimization}
After obtaining the optimized system prompt $x_s^*$ and the user prompt complement dataset $X_u^*$ (the optimization results of the final round in Sec \ref{sec:sys}), the online prediction process can be represented as:
\begin{equation}
    y=LLM(x_s^*, f(x_u|X^*_u))
\end{equation}
where $f$ is the online optimization function based on $X^*_u$. A common implementation of $f$, as adopted in BPO~\citep{BPO} and PAS~\citep{PAS}, involves fine-tuning a smaller language model on $X_u^*$. 

Here we propose an alternative implementation of $f$, in-context learning with the collected dataset (P3-ICL). For each input user query, we retrieve relevant user prompts along with their associated complementary instructions from $X^*_u$ to perform few-shot in-context learning. The prompt template is shown in Appendix~\ref{appendix:icl}. This approach mitigates the extra computational budget brought by the online optimization model while still achieving competitive performance gains. By leveraging retrieved demonstrations directly, we preserve inference efficiency without sacrificing the benefits of prompt optimization.

\section{Experiments}
\begin{table*}
  \vspace{-1em}
  \centering
  \normalsize
  \resizebox{1.0\linewidth}{!}{
    \setlength{\tabcolsep}{3mm}{
      \begin{tabular}{l|l|ccc>{\columncolor{gray!20}}c>{\columncolor{gray!20}}c}
        \hline
        \textbf{Dataset} & \textbf{Main Model} & \textbf{Raw} & \textbf{BPO} & \textbf{PAS} & \textbf{P3} & \textbf{P3-ICL}\\
        \hline
        
        \multirow{6}{*}{Arena-hard}           & GPT-4-turbo-2024-04-09  & 76.60 & 76.60 & 76.90 & \textbf{78.00} & 76.27\\
                                              & GPT-4-1106-preview      & 74.80 & 74.60 & \textbf{78.80} & 77.21 & 77.18\\
                                              & GPT-3.5-turbo-1106      & 18.90 & 15.90 & 22.10 & 25.56 & \textbf{36.81}\\
                                              & Qwen2-72b-Instruct      & 48.10 & 44.40 & 52.20 & 52.82 & \textbf{61.88} \\ 
                                              & LLaMA-3-70b-Instruct    & 41.10 & 45.20 & 50.30 & 51.64 & \textbf{51.96} \\ \cline{2-7}
                                              & \textbf{Average}        & 51.90 & 51.34 & 56.06 & 57.05 & \textbf{60.82} \\ 
        \hline
        \multirow{6}{*}{Alpaca-Eval 2.0}      & GPT-4-turbo-2024-04-09  & 46.12 & 54.65 & 65.31 & \textbf{70.50} & 67.70\\
                                              & GPT-4-1106-preview      & 50.00 & 55.19 & 65.92 & \textbf{69.94} & 66.40\\
                                              & GPT-3.5-turbo-1106      & 9.20  & 10.25 & 15.82 & 34.53 & \textbf{45.77}\\
                                              & Qwen2-72b-Instruct      & 31.70 & 31.25 & 45.53 & 61.37 & \textbf{64.84} \\ 
                                              & LLaMA-3-70b-Instruct    & 33.18 & 38.92 & 45.01 & 49.44 & \textbf{50.75} \\ \cline{2-7}
                                              & \textbf{Average}        & 34.04 & 38.05 & 47.52 & 57.16 & \textbf{59.09} \\
        \hline
        \multirow{6}{*}{Alpaca-Eval 2.0 (LC)} & GPT-4-turbo-2024-04-09  & 55.02 & 55.28 & 56.54 & \textbf{58.31} & 56.46\\
                                              & GPT-4-1106-preview      & 50.00 & 52.91 & 53.63 & \textbf{56.95} & 56.27\\
                                              & GPT-3.5-turbo-1106      & 19.30 & 20.29 & 23.31 & 35.35 & \textbf{42.20}\\
                                              & Qwen2-72b-Instruct      & 39.24 & 39.02 & 44.31 & \textbf{55.72} & 50.35 \\ 
                                              & LLaMA-3-70b-Instruct    & 34.42 & 39.24 & 40.52 & \textbf{42.15} & 42.10 \\ \cline{2-7}
                                              & \textbf{Average}        & 39.60 & 41.35 & 43.66 & \textbf{49.70} & 49.48 \\
        \hline
      \end{tabular}
      }
  }
  \caption{Main results on different prompt optimization methods for general QA. }
  \vspace{-0.5em}
  \label{tab:main}
\end{table*}

In this section, we conduct thorough experiments to analyze the performance of our proposed methods.  We report results with the user prompt optimization iteration number set to 1. Additional experiments demonstrating performance improvements with increased iteration numbers are provided in the Appendix~\ref{appendix:abl}. Other hyperparemeter settings are listed in Appendix~\ref{appendix:hyper}. Details of online model training are shown in Appendix~\ref{appendix:online}.

\subsection{Evaluation on General QA Task}
\label{exp:general}
We first evaluate our method on general question-answering scenarios. This domain serves as a critical testbed for prompt optimization frameworks due to its practical relevance to industrial LLM applications.

\subsubsection{Baselines and Settings}

\textbf{Benchmark:} In this part, we adopt three popular general QA benchmarks as in PAS: 

(1) \textbf{Arena-hard}~\citep{Arena-Hard}: A multi-domain benchmark testing model robustness through noise resilience, high-dimensional feature handling, and adversarial robustness. It evaluates both accuracy and generalization capabilities, including perturbation resistance and performance on unseen data.

(2) \textbf{Alpaca-Eval 2.0}~\citep{alpaca_eval}: A comprehensive benchmark extending its predecessor with diverse and challenging tasks to assess advanced LLMs across linguistic complexity, reasoning depth, and task variety. 

(3) \textbf{Alpaca-Eval 2.0 (LC)}~\citep{alpaca-eval-lc}: the length-controlled version of Alpaca-Eval 2.0. By introducing a regression-based debiasing strategy, it ensures that the evaluation of models is not skewed by the length of the responses. 
This version demonstrates stronger correlation with human evaluations, enabling fairer model comparisons.




\noindent \textbf{Baseline Methods}. To verify the advantages of P3 over other prompt optimization methods, we select two preceding online prompt optimization methods, PAS and BPO. These are two strong online prompt optimization baselines for improving human preference of LLMs' outputs in general scenarios. For reference, we also report the original results for LLMs without prompt optimization.

\noindent\textbf{Settings}. For these tasks, we use the dataset provided by PAS for offline optimization, ensuring a fair comparison. We select GPT4o-mini~\citep{GPT4o} as the proposal and optimizer model, and GPT-3.5-turbo-1106 as the base model for generating responses in the online stage. At the online optimization stage, we finetune a Qwen2-7B-instruct~\citep{Qwen2} model as the online optimization model. For P3-ICL, we use a light-weight embedding model~\citep{all-MiniLM-L6-v2} to retrieve data from the offline dataset. For benchmark evaluation, we choose the same base models following \cite{PAS} for fair comparisons: GPT-4-turbo-2024-04-09, GPT-4-1106-preview, GPT-3.5-turbo-1106, Qwen2-72b-Instruct, and LLaMA-3-70b-Instruct~\citep{llama3}. 


\subsubsection{Results}
As shown in Table~\ref{tab:main}, our methods demonstrate significant improvements across all evaluation dimensions. The base P3 framework dominates BPO and PAS in all but one model-dataset combination. Particularly notable are its gains on smaller models, e.g., +18.71\% over PAS for GPT-3.5-turbo on Alpaca-Eval 2.0. This validates the effectiveness of joint system-user prompt optimization. P3-ICL provides a competitive alternative and eliminates the need for dedicated model training. Remarkably, P3-ICL even surpasses P3 on weaker models like GPT-3.5-turbo and Qwen2, suggesting its particular value for resource-constrained deployments. Notably, P3 maintains its performance advantage even when using GPT-4o-mini as the optimizer, a smaller and weaker model than the GPT-4o employed by PAS. This consistent outperformance across diverse LLMs underscores the effectiveness of P3's joint optimization mechanism, despite being trained offline with only a single model.

\subsection{Evaluation on Reasoning Task}

We also conduct experiments on two complex reasoning tasks to examine P3's efficacy to help improve LLMs' reasoning ability. 

\subsubsection{Baselines and Settings}

\textbf{Benchmark:} In this domain, we choose two challenging benchmarks:

(1) \textbf{GSM8k}~\citep{cobbe2021gsm8k}: A dataset of linguistically diverse, high-quality math word problems at grade-school level. In this task, We optimize GPT-3.5-turbo using GPT-4o, achieving performance gains consistent with established methodologies~\citep{yuksekgonul2024textgrad, khattab2024dspy}.

(2) \textbf{GPQA}~\citep{rein2024gpqa}: A challenging benchmark containing multiple-choice questions in PhD-level science subjects. In this task, We deploy GPT-4o for self-optimization. During evaluation, iterative test-time updates (3 cycles) are aggregated via majority voting to determine final predictions.

\noindent\textbf{Baselines and Settings:} For both tasks, we compare the results of P3 and P3-ICL with popular APO methods, including 1) Zero-shot Chain-of-Thought (CoT), 2) vanilla in-context learning (ICL), 3) TextGrad~\citep{yuksekgonul2024textgrad}, 4) DSPy's BootstrappedFewShotRandomSearch (BFSR) optimizer~\citep{khattab2024dspy}, 5) PAS. In the offline optimization stage, unlike in Sec~\ref{exp:general}, we perform separate optimization for the two benchmarks, as they each provide training and evaluation datasets. For all the ICL methods, we apply 4-shot demonstrations.

\begin{table}[t]
  \centering
  \normalsize 
  \setlength{\tabcolsep}{4mm}{
      \begin{tabular}{c|cc}
        \hline
        \textbf{Methods} & \textbf{GSM8k}  & \textbf{GPQA}\\
        \hline
        zero-shot CoT  &  72.9 & 51.0\\
        few-shot CoT & 81.4 & 49.5\\
        TextGrad & 81.1 & 55.0\\
        DSPy & 81.1 & 50.5 \\
        PAS & 81.3 & 53.5\\
        \hline
        P3-ICL & 82.1 & 52.0 \\
        P3  &  \textbf{84.8} & \textbf{57.1}\\
        \hline
      \end{tabular}
      }
  \caption{Results on reasoning tasks.}
  \vspace{-0.5em}
  \label{tab:reasoning}
\end{table}

\subsubsection{Results}
The evaluation results are listed in Table~\ref{tab:reasoning}. P3 achieves marginal improvements against other methods. In GSM8k, P3 improves the performance of gpt-3.5-turbo-0125 from 72.9\% (vanilla CoT) to 84.8\%, outperforming other methods by at least 3.4\%. In GPQA, P3 also achieves promising results, which outperforms other baselines by at least 2.1\%. Besides, P3-ICL shows consistent improvements in both tasks against vanilla CoT and DSPy, indicating that introducing explicit reasoning instruction in demonstrations can benefit LLMs' reasoning performance.

\begin{table*}
  \vspace{0em}
  \centering
  \normalsize
  \resizebox{1.0\linewidth}{!}{
    \setlength{\tabcolsep}{3mm}{
      \begin{tabular}{llccccc}
        \hline
        \multirow{2}{*}{Main Model} & \multirow{2}{*}{\textbf{Methods}}  & \multirow{2}{*}{\textbf{AH}} & \multirow{2}{*}{\textbf{AE 2.0}} & \textbf{AE 2.0} & \multirow{2}{*}{\textbf{Avg.}} & \multirow{2}{*}{\textbf{$\triangle$}}\\
         &  &    &   &  (LC)  &   & \\   
        \hline
        
        \multirow{3}{*}{GPT-4-turbo-2024-04-09} &  P3  & 78.00 & 70.50 & 58.31 & 68.94 & \textbf{+9.69}\\
                               &  P3 w/o system  & 78.43 & 66.02 & 57.96 & 67.47 & +8.22\\
                               & PAS + system & 69.80 & 68.39 & 57.54 & 65.24 & +5.99\\
        \hline
        \multirow{3}{*}{GPT-4-1106-preview}     &  P3  & 77.21 & 69.94 & 56.95  & 68.03 & \textbf{+9.76}\\
                               &  P3 w/o system  & 77.87 & 65.86 & 56.92 & 66.88 & +8.61\\
                               & PAS + system & 72.25 & 65.03 & 53.07 & 63.45 & +5.18 \\
        \hline
        \multirow{3}{*}{GPT-3.5-turbo-1106}     &  P3  & 25.56 & 34.53 & 35.35 & 31.81 & \textbf{+16.01}\\
                               &  P3 w/o system  & 23.32 & 16.65 & 24.67 & 21.55 & +5.75\\
                               &  PAS + system  &  23.40  & 35.19 & 35.16 & 31.25 & +15.45 \\
        \hline
        \multirow{3}{*}{Qwen2-72b-Instruct}     &  P3  & 52.82  & 61.37 & 55.72 & 56.64 & \textbf{+16.96}\\
                               &  P3 w/o system  & 53.83 & 47.70 & 44.94 & 48.82 & +9.14\\
                               &  PAS + system  &  48.23  & 55.96 & 49.40 & 51.20 & +11.52\\
        \hline
        \multirow{3}{*}{LLaMA-3-70b-Instruct}   &  P3  & 51.64 & 49.44 & 42.15 & 47.74 & \textbf{+11.51}\\
                               &  P3 w/o system  & 50.85 & 47.14 & 40.76 & 46.25 & +10.02\\
                               &  PAS + system  &  47.26  & 46.52 & 39.86 & 44.55 & +8.32\\
        \hline
      \end{tabular}
    }
  }
  \caption{Ablation study on system prompt optimization in P3.}
    \vspace{-0.5em}
  \label{tab:ablation_system}
\end{table*}

\subsection{Ablation Study}

In this section, we conduct an ablation study to evaluate the impact of our design choices on LLM performance. Additionally, we analyze how these design choices address the three key challenges outlined earlier.

\subsubsection{Improved Affinity via System Prompt Optimization}
By jointly optimizing system and user prompts, P3 achieves enhanced affinity between the two components compared to previous methods.  We analyze the effect of system prompt optimization in this experiment. As shown in Table~\ref{tab:ablation_system}, P3 consistently outperforms its variant without system prompt optimization across all benchmarks and LLMs, demonstrating the stability and effectiveness of this approach. Notably, the improvements are more pronounced in prompt-sensitive LLMs, such as GPT-3.5-turbo and Qwen2. To further investigate the generalizability of this affinity, we integrate P3's system prompt into PAS. While PAS with P3's system prompt achieves remarkable gains in Alpaca-Eval and Alpaca-Eval (LC), its performance degrades badly in Arena Hard, underscoring the necessity of joint optimization for robust performance.

The benefits of joint system prompt optimization are twofold. First, it leverages task-specific information to steer LLMs toward task-aligned behaviors. Second, it enhances the model's ability to adhere to the details provided in complementary instructions, ensuring more precise and context-aware responses. We provide a comprehensive case study in Appendix~\ref{appendix:case}.
And all the optimized system prompts are listed in Appendix~\ref{appendix:opt}.

\subsubsection{Enhanced Diversity in User Prompt Complementary}

\begin{figure}[tb]
\hspace{-1em}
\centering
  \includegraphics[width=0.5\textwidth]{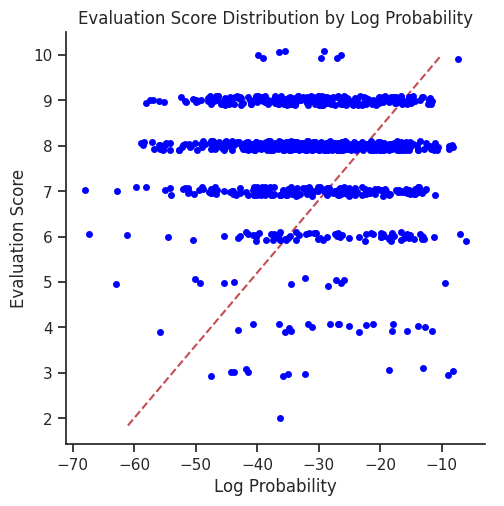}
   \vspace{-1em}
  \caption {Plots for complements' evaluation score by the generation log probability.}
  \vspace{-0.5em}
\label{fig:logprob}
\end{figure}

The potential complements for each user prompt typically result in multiple diverse outcomes. Previous methods, such as PAS and BPO, focus exclusively on a singular complement or revision for each user prompt, thus limiting exploration in the massive solution space. That is because greedy decoding does not necessarily lead to the best complements. As shown in Fig.~\ref{fig:logprob}, there's no positive correlation between log probability and eval score. Thus, high log probability during generation doesn't guarantee a high score, demonstrating the importance of diversity sampling in the complementary instruction optimization process.  

P3 mitigates this limitation through diverse sampling and generation. As illustrated in Table \ref{tab:ablation_diversity}, the results of P3 without system prompt demonstrate superior performance compared with PAS. This evinces that P3's diverse sampling and generation strategy effectively enhances overall performance.

\begin{table}[t]
\normalsize
\resizebox{\linewidth}{!}{
    \setlength{\tabcolsep}{4mm}{
    \begin{tabular}{lll}
    \toprule
    \textbf{Dataset} & \textbf{Method}   & \textbf{Score}\\ \midrule
    \multirow{2}{*}{Arena Hard} & PAS & 22.10 \\ &  P3 w/o system & \textbf{23.32}\\
    \midrule
    Alpaca-Eval & PAS & 15.82 \\ 2.0 & P3 w/o system & \textbf{16.65}\\
   \midrule
    Alpaca-Eval &  PAS & 23.32 \\ 2.0 (LC) & P3 w/o system & \textbf{24.67}\\
    \midrule
    \multirow{2}{*}{GSM8k} & PAS & 81.30 \\ & P3 w/o system & \textbf{83.10}\\
    \bottomrule
    \end{tabular}
    }
    }
    \centering
    \caption{Ablation study for diverse generation. All results are generated by GPT-3.5-turbo-1106.}
    \vspace{-1em}
    \label{tab:ablation_diversity}
\end{table}

\subsubsection{Inference Efficiency via In-Context Learning}

\begin{table}[t]
  \centering
  \normalsize
  \resizebox{\linewidth}{!}{
    \begin{tabular}{l|ccc}
      \hline
      \textbf{Methods} & \textbf{PAS} & \textbf{P3} & \textbf{P3-ICL} \\
      \hline
      Param. (Billion) & 7.62 & 7.62 & 0.02 \\
      Mem. (MiB) & $18000$& $18000$& $5000$\\
      Extra Lat. (s) & 0.53 & 0.53 & 0.07 \\
      FTL (s) & 2.01 & 2.03 & 1.46 \\
      Overall Lat. (s) & 13.86 & 13.89 & 13.40 \\
      \hline
      \end{tabular}
  }
  \caption{The computational complexity comparison between PAS, P3, and P3-ICL. 1) \textit{Param} represents the model parameters used in the online optimization stage. 2) \textit{Mem} represents the memory occupancy in the online optimization period. 3) \textit{Extra Lat.} represents the average extra latency brought by prompt optimization. 4) \textit{FTL} represets the average first-token latency and 5) \textit{Overall Lat.} represents the average overall latency.}
  \vspace{-1em}
  \label{tab:compute_complexity}
\end{table}

The offline constructed data can not only be used for training a small language model to conduct online optimization, but also for online in-context learning, considering the efficiency of the inference stage. Table~\ref{tab:ablation_icl} shows the ablation results for P3-ICL, which shows its superior performance against vanilla ICL. Considering the results from Table~\ref{tab:main}, we can further conclude that P3-ICL can still outperform PAS and BPO across all benchmarks, and gain stable improvements against vanilla ICL. We can conclude that the offline constructed data is also compatible with in-context learning and effective in guiding LLMs to produce better responses.

\begin{table*}
  \vspace{-1em}
  \centering
  \normalsize
  \resizebox{1.0\linewidth}{!}{
    \setlength{\tabcolsep}{3.5mm}{
      \begin{tabular}{llccccc}
        \hline
        \multirow{2}{*}{Main Model} & \multirow{2}{*}{\textbf{Methods}}  & \multirow{2}{*}{\textbf{AH}} & \multirow{2}{*}{\textbf{AE 2.0}} & \textbf{AE 2.0} & \multirow{2}{*}{\textbf{Avg.}} & \multirow{2}{*}{\textbf{$\triangle$}}\\
        &  &    &   &  (LC)  &   & \\   
        \hline
        \multirow{2}{*}{GPT-4-turbo-2024-04-09} 
                               & P3-ICL    & 76.27 & 67.70 & 56.46 & 64.92 & \textbf{+5.67}\\
                               & ICL       & 74.02 & 64.97 & 56.94 &  63.80 & +4.55\\
        \hline
        \multirow{2}{*}{GPT-4-1106-preview}     
                               &  P3-ICL    & 77.18 & 66.40 & 56.27 & 64.53 & +6.26\\
                               &  ICL       & 76.89 & 66.02 & 58.12 & 64.83 & \textbf{+6.56}\\
        \hline
        \multirow{2}{*}{GPT-3.5-turbo-1106}     
                               &  P3-ICL    & 36.81 & 45.77 & 42.20 & 41.59 & \textbf{+25.79}\\
                               &  ICL       & 31.65 & 28.23 & 35.80 & 31.89 & +16.09\\
        \hline
        \multirow{2}{*}{Qwen2-72b-Instruct}     
                               &  P3-ICL    & 61.88 & 64.84 & 50.35 & 59.02 & \textbf{+19.34}\\
                               &  ICL       & 60.57 & 57.58 & 49.88 & 56.01 & +16.33\\
        \hline
        \multirow{2}{*}{LLaMA-3-70b-Instruct}  
                               &  P3-ICL    & 51.96 & 50.75  & 42.10 & 48.27 & \textbf{+12.04}\\
                               &  ICL       & 49.80 & 45.60 & 41.08 & 45.49 & +9.26\\
        \hline

      \end{tabular}
    }
  }
  \caption{Ablation study for P3-ICL. We show the superior performance of P3-ICL against vanilla ICL.}
  \vspace{-0.5em}
  \label{tab:ablation_icl}
\end{table*}

The core advantage of P3-ICL is more efficient and economic inference. As shown in Table~\ref{tab:compute_complexity}, P3-ICL reduces the memory occupancy from 18000 MiB to 5000 MiB compared with using the online model for query complement, and shortens the extra latency from 530 ms to 70 ms and offers a 25\% lower first-token latency(on a single 3090 RTS GPU). Crucially, we observed that the longer input prompts generated by P3-ICL did not lead to a significant increase in overall latency. The reason is that P3-ICL only increases the number of tokens processed during the \textbf{prefill} phase, where the efficiency of parallel processing on modern hardware can mitigate its impact on total inference time. Thus, the overall latency is usually dominated by the time spent generating tokens in the \textbf{decode} phase, where P3 and PAS will bring larger latency with their generated complements.

From the above observations, we can conclude that the inference efficiency of P3 can be substantially improved through in-context learning while still achieving comparable performance with previous methods.

\section{Related Work}

\subsection{Prompt Optimization}
Prompt optimization is crucial for enhancing the performance of pretrained language models~\citep{Bert,T5,GPT}, as it allows for more accurate and contextually relevant outputs through revision the input prompt. Traditional methods of prompt optimization, such as soft prompt tuning~\citep{prefix-tuning, PEPT}, have shown significant improvements in model performance by adjusting the prompts during the training phase. Recently, hard prompt optimizations have gained traction, where directly modifying the discrete prompts can leverage the efficacy of pretrained models~\cite{GPT3,multitaskPrompt,instruct_tune}.

\subsection{Automatic Prompt Optimization}
Automatic prompt optimization~\citep{Autoprompt} aims to develop algorithms that can autonomously improve prompts without human efforts. For instance, \citet{Rlprompt} utilizes reinforcement learning to dynamically adjust prompts. \citet{EPO} employs evolutionary algorithms to iteratively generate and refine prompts. \citet{yuksekgonul2024textgrad} and \citet{cheng2024trace} perform end-to-end optimization for agentic system with feedback proposed by LLMs, analogy to AutoDiff in deep learning system. These automated techniques enhance the scalability and adaptability of prompt optimization across diverse domains.

\subsection{Query-Dependent Automatic Prompt Optimization}
Query-dependent prompt optimization tailors prompt optimization to the specific queries. This approach recognizes that a one-size-fits-all prompt may not be effective for diverse queries and seeks to adapt prompts based on the input queries. \citet{BPO} proposed a method where the prompt is dynamically adjusted according to the query characteristics, leading to more precise and contextually relevant responses. Additionally, \citet{PAS} highlighted the risk of losing the original meaning during online prompt modifications and developed a framework that generates context-aware prompt complements, thereby improving the overall performance. These query-dependent approaches highlight the importance of contextual adaptability in prompt optimization, ensuring precise and targeted optimization for a wide range of user inputs. Our work falls within this category.

\section{Conclusion}
In this work, we introduce Prompts Promote Prompting (P3), a novel framwork addressing three fundamental challenges in online prompt optimization: affinity alignment, response diversity, and inference efficiency. Our solution establishes a dual-phase optimization paradigm that combines offline prompt optimization with online adaptation of user prompts. By optimizing system prompts offline and dynamically adjusting user prompts online, P3 ensures the synergistic function of the two core components in prompt engineering. P3-ICL, the in-context learning variant of P3, achieves better efficiency in online stage. Experimental results show that P3 and P3-ICL significantly outperform existing automatic prompt optimization methods such as BPO and PAS across various LLMs and domains. This approach provides a more flexible and efficient solution for prompt optimization, providing a promising way for enhanced LLM applications in real-time scenarios.

\section{Limitations}
We identify that there may be some possible limitations in this research. Firstly, due to the limit of resources, we only choose Qwen2-7B-Instruct as the base model for online optimization. We believe using models of similar (e.g., Llama-3-8B~\cite{meta2024llama3}) or even smaller size (e.g., Qwen2.5-3B~\cite{qwen25technicalreport}) will reach comparable performance. We will conduct further experiments to validate this point. Secondly, our experiments focus on standard instruction-tuned LLMs, aligning with other APO methods. We will investigate the potential of leveraging P3 to enhance emerged reasoning models (e.g., OpenAI O1~\citep{jaech2024openai}, DeepSeek-R1~\citep{guo2025deepseek}) in future research.

\bibliography{custom}

\begin{thebibliography}{39}
\providecommand{\natexlab}[1]{#1}

\bibitem[{Brown et~al.(2020)Brown, Mann, Ryder, Subbiah, Kaplan, Dhariwal, Neelakantan, Shyam, Sastry, Askell et~al.}]{GPT3}
Tom Brown, Benjamin Mann, Nick Ryder, Melanie Subbiah, Jared~D Kaplan, Prafulla Dhariwal, Arvind Neelakantan, Pranav Shyam, Girish Sastry, Amanda Askell, et~al. 2020.
\newblock Language models are few-shot learners.
\newblock \emph{Advances in neural information processing systems}, 33:1877--1901.

\bibitem[{Cheng et~al.(2024)Cheng, Nie, and Swaminathan}]{cheng2024trace}
Ching-An Cheng, Allen Nie, and Adith Swaminathan. 2024.
\newblock \href {https://openreview.net/forum?id=rYs2Dmn9tD} {Trace is the next autodiff: Generative optimization with rich feedback, execution traces, and {LLM}s}.
\newblock In \emph{The Thirty-eighth Annual Conference on Neural Information Processing Systems}.

\bibitem[{Cheng et~al.(2023)Cheng, Liu, Zheng, Ke, Wang, Dong, Tang, and Huang}]{BPO}
Jiale Cheng, Xiao Liu, Kehan Zheng, Pei Ke, Hongning Wang, Yuxiao Dong, Jie Tang, and Minlie Huang. 2023.
\newblock Black-box prompt optimization: Aligning large language models without model training.
\newblock \emph{arXiv preprint arXiv:2311.04155}.

\bibitem[{Chung et~al.(2024)Chung, Hou, Longpre, Zoph, Tay, Fedus, Li, Wang, Dehghani, Brahma et~al.}]{instruct_tune}
Hyung~Won Chung, Le~Hou, Shayne Longpre, Barret Zoph, Yi~Tay, William Fedus, Yunxuan Li, Xuezhi Wang, Mostafa Dehghani, Siddhartha Brahma, et~al. 2024.
\newblock Scaling instruction-finetuned language models.
\newblock \emph{Journal of Machine Learning Research}, 25(70):1--53.

\bibitem[{Cobbe et~al.(2021)Cobbe, Kosaraju, Bavarian, Chen, Jun, Kaiser, Plappert, Tworek, Hilton, Nakano, Hesse, and Schulman}]{cobbe2021gsm8k}
Karl Cobbe, Vineet Kosaraju, Mohammad Bavarian, Mark Chen, Heewoo Jun, Lukasz Kaiser, Matthias Plappert, Jerry Tworek, Jacob Hilton, Reiichiro Nakano, Christopher Hesse, and John Schulman. 2021.
\newblock Training verifiers to solve math word problems.
\newblock \emph{arXiv preprint arXiv:2110.14168}.

\bibitem[{Deng et~al.(2022)Deng, Wang, Hsieh, Wang, Guo, Shu, Song, Xing, and Hu}]{Rlprompt}
Mingkai Deng, Jianyu Wang, Cheng-Ping Hsieh, Yihan Wang, Han Guo, Tianmin Shu, Meng Song, Eric~P Xing, and Zhiting Hu. 2022.
\newblock Rlprompt: Optimizing discrete text prompts with reinforcement learning.
\newblock \emph{arXiv preprint arXiv:2205.12548}.

\bibitem[{Devlin et~al.(2019)Devlin, Chang, Lee, and Toutanova}]{Bert}
Jacob Devlin, Ming-Wei Chang, Kenton Lee, and Kristina Toutanova. 2019.
\newblock Bert: Pre-training of deep bidirectional transformers for language understanding.
\newblock In \emph{Proceedings of the 2019 Conference of the North American Chapter of the Association for Computational Linguistics: Human Language Technologies, Volume 1 (Long and Short Papers)}, pages 4171--4186.

\bibitem[{Dubey et~al.(2024)Dubey, Jauhri, Pandey, Kadian, Al-Dahle, Letman, Mathur, Schelten, Yang, Fan et~al.}]{llama3}
Abhimanyu Dubey, Abhinav Jauhri, Abhinav Pandey, Abhishek Kadian, Ahmad Al-Dahle, Aiesha Letman, Akhil Mathur, Alan Schelten, Amy Yang, Angela Fan, et~al. 2024.
\newblock \href {https://arxiv.org/abs/2407.21783} {The llama 3 herd of models}.
\newblock \emph{Preprint}, arXiv:2407.21783.

\bibitem[{Dubois et~al.(2024)Dubois, Galambosi, Liang, and Hashimoto}]{alpaca-eval-lc}
Yann Dubois, Bal{\'a}zs Galambosi, Percy Liang, and Tatsunori~B Hashimoto. 2024.
\newblock Length-controlled alpacaeval: A simple way to debias automatic evaluators.
\newblock \emph{arXiv preprint arXiv:2404.04475}.

\bibitem[{Fernando et~al.()Fernando, Banarse, Michalewski, Osindero, and Rockt{\"a}schel}]{Promptbreeder}
Chrisantha Fernando, Dylan~Sunil Banarse, Henryk Michalewski, Simon Osindero, and Tim Rockt{\"a}schel.
\newblock Promptbreeder: Self-referential self-improvement via prompt evolution.
\newblock In \emph{Forty-first International Conference on Machine Learning}.

\bibitem[{Guo et~al.(2025)Guo, Yang, Zhang, Song, Zhang, Xu, Zhu, Ma, Wang, Bi et~al.}]{guo2025deepseek}
Daya Guo, Dejian Yang, Haowei Zhang, Junxiao Song, Ruoyu Zhang, Runxin Xu, Qihao Zhu, Shirong Ma, Peiyi Wang, Xiao Bi, et~al. 2025.
\newblock Deepseek-r1: Incentivizing reasoning capability in llms via reinforcement learning.
\newblock \emph{arXiv preprint arXiv:2501.12948}.

\bibitem[{Guo et~al.()Guo, Wang, Guo, Li, Song, Tan, Liu, Bian, and Yang}]{EPO}
Qingyan Guo, Rui Wang, Junliang Guo, Bei Li, Kaitao Song, Xu~Tan, Guoqing Liu, Jiang Bian, and Yujiu Yang.
\newblock Connecting large language models with evolutionary algorithms yields powerful prompt optimizers.
\newblock In \emph{The Twelfth International Conference on Learning Representations}.

\bibitem[{Hayou et~al.(2024)Hayou, Ghosh, and Yu}]{hayou2024loraplus}
Soufiane Hayou, Nikhil Ghosh, and Bin Yu. 2024.
\newblock Lora+: Efficient low rank adaptation of large models.
\newblock \emph{arXiv 2402.12354}.

\bibitem[{Jaech et~al.(2024)Jaech, Kalai, Lerer, Richardson, El-Kishky, Low, Helyar, Madry, Beutel, Carney et~al.}]{jaech2024openai}
Aaron Jaech, Adam Kalai, Adam Lerer, Adam Richardson, Ahmed El-Kishky, Aiden Low, Alec Helyar, Aleksander Madry, Alex Beutel, Alex Carney, et~al. 2024.
\newblock Openai o1 system card.
\newblock \emph{arXiv preprint arXiv:2412.16720}.

\bibitem[{Khattab et~al.(2024)Khattab, Singhvi, Maheshwari, Zhang, Santhanam, Vardhamanan, Haq, Sharma, Joshi, Moazam, Miller, Zaharia, and Potts}]{khattab2024dspy}
Omar Khattab, Arnav Singhvi, Paridhi Maheshwari, Zhiyuan Zhang, Keshav Santhanam, Sri Vardhamanan, Saiful Haq, Ashutosh Sharma, Thomas~T. Joshi, Hanna Moazam, Heather Miller, Matei Zaharia, and Christopher Potts. 2024.
\newblock Dspy: Compiling declarative language model calls into self-improving pipelines.

\bibitem[{Khattab et~al.(2023)Khattab, Singhvi, Maheshwari, Zhang, Santhanam, Vardhamanan, Haq, Sharma, Joshi, Moazam et~al.}]{Dspy}
Omar Khattab, Arnav Singhvi, Paridhi Maheshwari, Zhiyuan Zhang, Keshav Santhanam, Sri Vardhamanan, Saiful Haq, Ashutosh Sharma, Thomas~T Joshi, Hanna Moazam, et~al. 2023.
\newblock Dspy: Compiling declarative language model calls into self-improving pipelines.
\newblock \emph{arXiv preprint arXiv:2310.03714}.

\bibitem[{Lester et~al.(2021)Lester, Al-Rfou, and Constant}]{PEPT}
Brian Lester, Rami Al-Rfou, and Noah Constant. 2021.
\newblock The power of scale for parameter-efficient prompt tuning.
\newblock In \emph{Proceedings of the 2021 Conference on Empirical Methods in Natural Language Processing}, pages 3045--3059.

\bibitem[{Li et~al.(2024)Li, Chiang, Frick, Dunlap, Wu, Zhu, Gonzalez, and Stoica}]{Arena-Hard}
Tianle Li, Wei-Lin Chiang, Evan Frick, Lisa Dunlap, Tianhao Wu, Banghua Zhu, Joseph~E Gonzalez, and Ion Stoica. 2024.
\newblock From crowdsourced data to high-quality benchmarks: Arena-hard and benchbuilder pipeline.
\newblock \emph{arXiv preprint arXiv:2406.11939}.

\bibitem[{Li and Liang(2021)}]{prefix-tuning}
Xiang~Lisa Li and Percy Liang. 2021.
\newblock Prefix-tuning: Optimizing continuous prompts for generation.
\newblock \emph{arXiv preprint arXiv:2101.00190}.

\bibitem[{Li et~al.(2023)Li, Zhang, Dubois, Taori, Gulrajani, Guestrin, Liang, and Hashimoto}]{alpaca_eval}
Xuechen Li, Tianyi Zhang, Yann Dubois, Rohan Taori, Ishaan Gulrajani, Carlos Guestrin, Percy Liang, and Tatsunori~B. Hashimoto. 2023.
\newblock Alpacaeval: An automatic evaluator of instruction-following models.
\newblock \url{https://github.com/tatsu-lab/alpaca_eval}.

\bibitem[{Meta(2024)}]{meta2024llama3}
Meta. 2024.
\newblock \href {https://arxiv.org/abs/2407.21783} {The llama 3 herd of models}.
\newblock \emph{Preprint}, arXiv:2407.21783.

\bibitem[{OpenAI(2024{\natexlab{a}})}]{openai2024gpt4}
OpenAI. 2024{\natexlab{a}}.
\newblock \href {https://arxiv.org/abs/2303.08774} {Gpt-4 technical report}.
\newblock \emph{Preprint}, arXiv:2303.08774.

\bibitem[{OpenAI(2024{\natexlab{b}})}]{gpt-4o}
OpenAI. 2024{\natexlab{b}}.
\newblock gpt-4o.
\newblock \url{https://openai.com/index/gpt-4o-system-card/}.

\bibitem[{OpenAI(2024{\natexlab{c}})}]{GPT4o}
OpenAI. 2024{\natexlab{c}}.
\newblock Gpt4o.
\newblock \url{https://platform.openai.com/docs/models/gpt-4o}.

\bibitem[{Pryzant et~al.(2023)Pryzant, Iter, Li, Lee, Zhu, and Zeng}]{APO}
Reid Pryzant, Dan Iter, Jerry Li, Yin Lee, Chenguang Zhu, and Michael Zeng. 2023.
\newblock Automatic prompt optimization with “gradient descent” and beam search.
\newblock In \emph{Proceedings of the 2023 Conference on Empirical Methods in Natural Language Processing}, pages 7957--7968.

\bibitem[{Qwen(2024)}]{yang2024qwen2}
Qwen. 2024.
\newblock \href {https://arxiv.org/abs/2407.10671} {Qwen2 technical report}.
\newblock \emph{Preprint}, arXiv:2407.10671.

\bibitem[{Qwen(2025)}]{qwen25technicalreport}
Qwen. 2025.
\newblock \href {https://arxiv.org/abs/2412.15115} {Qwen2.5 technical report}.
\newblock \emph{Preprint}, arXiv:2412.15115.

\bibitem[{Radford et~al.()Radford, Narasimhan, Salimans, and Sutskever}]{GPT}
Alec Radford, Karthik Narasimhan, Tim Salimans, and Ilya Sutskever.
\newblock Improving language understanding by generative pre-training.

\bibitem[{Raffel et~al.(2020)Raffel, Shazeer, Roberts, Lee, Narang, Matena, Zhou, Li, and Liu}]{T5}
Colin Raffel, Noam Shazeer, Adam Roberts, Katherine Lee, Sharan Narang, Michael Matena, Yanqi Zhou, Wei Li, and Peter~J Liu. 2020.
\newblock Exploring the limits of transfer learning with a unified text-to-text transformer.
\newblock \emph{Journal of machine learning research}, 21(140):1--67.

\bibitem[{Rajbhandari et~al.(2020)Rajbhandari, Rasley, Ruwase, and He}]{rajbhandari2020zero}
Samyam Rajbhandari, Jeff Rasley, Olatunji Ruwase, and Yuxiong He. 2020.
\newblock Zero: Memory optimizations toward training trillion parameter models.
\newblock In \emph{SC20: International Conference for High Performance Computing, Networking, Storage and Analysis}, pages 1--16. IEEE.

\bibitem[{Rein et~al.(2024)Rein, Hou, Stickland, Petty, Pang, Dirani, Michael, and Bowman}]{rein2024gpqa}
David Rein, Betty~Li Hou, Asa~Cooper Stickland, Jackson Petty, Richard~Yuanzhe Pang, Julien Dirani, Julian Michael, and Samuel~R. Bowman. 2024.
\newblock \href {https://openreview.net/forum?id=Ti67584b98} {{GPQA}: A graduate-level google-proof q\&a benchmark}.
\newblock In \emph{First Conference on Language Modeling}.

\bibitem[{Sanh et~al.(2021)Sanh, Webson, Raffel, Bach, Sutawika, Alyafeai, Chaffin, Stiegler, Scao, Raja et~al.}]{multitaskPrompt}
Victor Sanh, Albert Webson, Colin Raffel, Stephen~H Bach, Lintang Sutawika, Zaid Alyafeai, Antoine Chaffin, Arnaud Stiegler, Teven~Le Scao, Arun Raja, et~al. 2021.
\newblock Multitask prompted training enables zero-shot task generalization.
\newblock \emph{arXiv preprint arXiv:2110.08207}.

\bibitem[{SBERT(2021)}]{all-MiniLM-L6-v2}
SBERT. 2021.
\newblock all-minilm-l6-v2.
\newblock \url{https://huggingface.co/sentence-transformers/all-MiniLM-L6-v2}.

\bibitem[{Shin et~al.(2020)Shin, Razeghi, Logan~IV, Wallace, and Singh}]{Autoprompt}
Taylor Shin, Yasaman Razeghi, Robert~L Logan~IV, Eric Wallace, and Sameer Singh. 2020.
\newblock Autoprompt: Eliciting knowledge from language models with automatically generated prompts.
\newblock \emph{arXiv preprint arXiv:2010.15980}.

\bibitem[{Yang et~al.(2024{\natexlab{a}})Yang, Yang, Hui, Zheng, Yu, Zhou, Li, Li, Liu, Huang, Dong, Wei, Lin, Tang et~al.}]{Qwen2}
An~Yang, Baosong Yang, Binyuan Hui, Bo~Zheng, Bowen Yu, Chang Zhou, Chengpeng Li, Chengyuan Li, Dayiheng Liu, Fei Huang, Guanting Dong, Haoran Wei, Huan Lin, Jialong Tang, et~al. 2024{\natexlab{a}}.
\newblock \href {https://arxiv.org/abs/2407.10671} {Qwen2 technical report}.
\newblock \emph{Preprint}, arXiv:2407.10671.

\bibitem[{Yang et~al.(2024{\natexlab{b}})Yang, Wang, Lu, Liu, Le, Zhou, and Chen}]{opro}
Chengrun Yang, Xuezhi Wang, Yifeng Lu, Hanxiao Liu, Quoc~V Le, Denny Zhou, and Xinyun Chen. 2024{\natexlab{b}}.
\newblock \href {https://openreview.net/forum?id=Bb4VGOWELI} {Large language models as optimizers}.
\newblock In \emph{The Twelfth International Conference on Learning Representations}.

\bibitem[{Yuksekgonul et~al.(2024)Yuksekgonul, Bianchi, Boen, Liu, Huang, Guestrin, and Zou}]{yuksekgonul2024textgrad}
Mert Yuksekgonul, Federico Bianchi, Joseph Boen, Sheng Liu, Zhi Huang, Carlos Guestrin, and James Zou. 2024.
\newblock \href {https://arxiv.org/abs/2406.07496} {Textgrad: Automatic "differentiation" via text}.

\bibitem[{Zheng et~al.(2024)Zheng, Liang, Yang, Sun, Li, Xiong, Zhang, Wu, Li, Sheng et~al.}]{PAS}
Miao Zheng, Hao Liang, Fan Yang, Haoze Sun, Tianpeng Li, Lingchu Xiong, Yan Zhang, Yozhen Wu, Kun Li, Yanjun Sheng, et~al. 2024.
\newblock {PAS}: Data-efficient plug-and-play prompt augmentation system.
\newblock \emph{arXiv preprint arXiv:2407.06027}.

\bibitem[{Zhou et~al.()Zhou, Muresanu, Han, Paster, Pitis, Chan, and Ba}]{APE}
Yongchao Zhou, Andrei~Ioan Muresanu, Ziwen Han, Keiran Paster, Silviu Pitis, Harris Chan, and Jimmy Ba.
\newblock Large language models are human-level prompt engineers.
\newblock In \emph{The Eleventh International Conference on Learning Representations}.

\end{thebibliography}

\clearpage

\section{Appendix}
\subsection{Ablation study on the search depths}
\label{appendix:abl}
\begin{figure}[H]
\centering
  \includegraphics[width=0.48\textwidth]{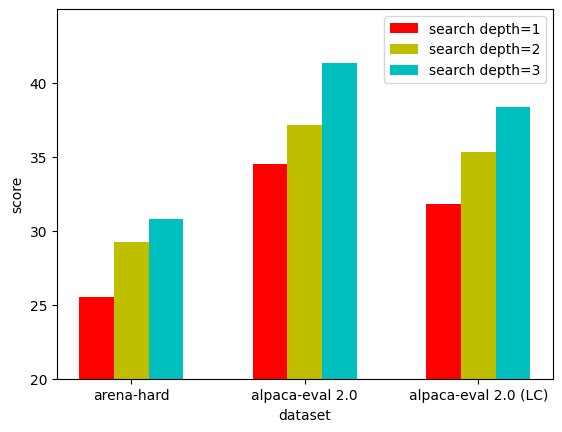}
  \caption {Effect of different search depths}
\label{fig:ablation_depth}
\end{figure}
In this ablation, we examine whether increasing the search depths will improve P3's end-to-end performance. We choose GPT-3.5-turbo-1106 as the base model, and increase the search depth (number of iterations) from 1 to 3. The evaluation is based on the Arena-hard and Alpaca-eval benchmarks. As illustrated in Fig.~\ref{fig:ablation_depth}, evaluation scores increase with search depth across all three benchmarks, justifying that P3 can generate more robust complementary instructions when searching more adequately on the solution space, thus improving overall performance.

\subsection{Optimized System prompts}
\label{appendix:opt}
Here we list the optimized system prompts in the offline optimization stage for all datasets. Notice that in GSM8k, we manfully include a piece of instruction to control the output format for evaluation.

\subsubsection{Optimized system prompt for general QA}
\textbf{\textcolor{red}{Initial prompt:}} \textsf{You are a helpful assistant. Given a user query, please consider the hints in the complementary instruction and provide a comprehensive, step-by-step solution.}

\noindent\textbf{\textcolor{red}{Optimized prompt:}} \textsf{As an intelligent assistant, your mission is to meticulously analyze user queries alongside their complementary instructions. Deliver a comprehensive, step-by-step response that not only meets the user's specific needs but also deepens their understanding of the topic. Ensure clarity, creativity, and relevance in your answers, effectively utilizing insights from the provided hints to create engaging and informative content across a wide range of subjects.}

\subsubsection{Optimized system prompt for GSM8k}
(The text in blue is added manually for output format control.)\\

\noindent\textbf{\textcolor{red}{Initial prompt:}} \textsf{You are an expert in solving math problems. Given a user query, please consider the hints in the complementary instruction and provide a comprehensive, step-by-step solution. \textcolor{blue}{Always conclude the last line of your response should be of the following format: "Answer: \$VALUE" where VALUE is a numerical value.}}

\noindent\textbf{\textcolor{red}{Optimized prompt:}} \textsf{You are a mathematics expert tasked with solving a variety of math problems. Upon receiving a user query, carefully analyze the provided hints in the complementary instruction to guide your solution process. Offer a detailed, step-by-step explanation of your approach, ensuring clarity and coherence throughout. \textcolor{blue}{Always conclude the last line of your response should be of the following format: "Answer: \$VALUE" where VALUE is a numerical value.}}

\subsubsection{Optimized system prompt for GPQA}
\textbf{\textcolor{red}{Initial prompt:}} \textsf{You are an expert in solving STEM problems. Given a user query, please consider the hints in the complementary instruction and provide a comprehensive, step-by-step solution.}

\noindent\textbf{\textcolor{red}{Optimized prompt:}} \textsf{As an expert in STEM problem-solving, your task is to provide a comprehensive and structured solution to the user's query. Begin by carefully analyzing the complementary hints provided in the user's instructions. Your response should be a detailed, step-by-step explanation that not only addresses the specific problem but also integrates relevant concepts and methodologies. Strive for clarity and depth in your explanations, while also encouraging creative approaches or alternative solutions where applicable. Ensure that your response is tailored to the user's needs, making it both informative and engaging.}

\clearpage

\onecolumn
\subsection{Prompt Template for ICL and P3-ICL}
\label{appendix:icl}

\begin{figure*}[ht]
\begin{tcbitemize}[raster columns=2, raster equal height]
\tcbitem[title={Prompt template for ICL},colback=yellow!10!white,colframe=red!75!black]
Given a user query, please follow the examples below to give a step-by-step and comprehensive answer. \\\\
\#\# Examples \\
User query: \\
\{example query 1\} \\\\
Answer: \\
\{example answer 1\} \\
... \\\\
User query: \\
\{example query N\} \\\\
Answer: \\
\{example answer N\} \\\\
\#\# Task \\
User query: \\
\{query\} \\\\
Answer:
\tcbitem[title={Prompt template for P3-ICL},colback=yellow!10!white,colframe=red!75!black]
Given a user query, please follow the examples below to give a step-by-step and comprehensive answer. \\\\
\#\# Examples \\
User query: \\
\{example query 1\} \\\\
Answer: \\
<complementary\_instruction> \\
\{example complement 1\} \\
</complementary\_instruction> \\
<answer> \\
\{example answer 1\} \\
</answer> \\
... \\\\
User query: \\
\{example query N\} \\\\
Answer: \\
<complementary\_instruction> \\
\{example complement N\} \\
</complementary\_instruction> \\
<answer> \\
\{example answer N\} \\
</answer> \\\\
\#\# Task \\
User query: \\
\{query\} \\\\
Answer:
\end{tcbitemize}
\end{figure*}

\vspace{0em}
\subsection{Case Study}
\label{appendix:case}
To better showcase the effectiveness of P3, this section provides a case study on how the affinity between the system prompt and user prompt affects the LLMs' performance.

In Fig.~\ref{fig:case_affinity}, we show the affinity of the jointly optimized system prompt for guiding LLMs to better adhere to the details of the complementary instruction. The query is a question from GSM8k dataset. For PAS, we employ a predefined system prompt with zero-shot CoT prompting and structured output control. For P3, we utilize the jointly optimized system prompt. PAS and P3 both output correct complementary instructions: calculate the net profit by subtracting the total cost from the total revenue. However, analysis of the LLM's response generated with PAS reveals a deviation from its complementary instruction. The LLM initially calculates the total number of candles that can be made, rather than the total revenue as specified in the instruction. As a result, it got the final answer wrong. Conversely, the LLM's answer generated with P3 precisely follows P3's complementary instruction, which leads to a correct answer.

\begin{figure}[H]
\centering
\resizebox{1.0\textwidth}{!}{
\begin{tcolorbox}[colback=blue!5!white,colframe=blue!75!black]
  \textbf{PAS system prompt}: You will answer a mathematical reasoning question. Think step by step. Always conclude the last line of your response should be of the following format: "Answer: \$VALUE" where VALUE is a numerical value. \\
  \textbf{PAS user prompt}: Charlie wants to sell beeswax candles. For every pound of beeswax, he can make 10 tapered candles. One pound of beeswax and the wicks cost \$10.00 in supplies. If he sells each candle for \$2.00 each, what is his net profit if he makes and sells 20 candles? \\ 
  \textcolor{blue}{To calculate Charlie's net profit, determine the total revenue from selling 20 candles and subtract the total cost of materials and wicks used in production. Ensure to use clear formulas and step-by-step calculations for accuracy.} 
  
  \textbf{PAS answer}: First, let's calculate the total number of candles Charlie can make with 1 pound of beeswax: 1 pound of beeswax * 10 candles/pound = 10 candles. \\
  Next, let's calculate the total cost of materials for making 20 candles: \$10.00. \\ 
  Now, let's calculate the total revenue from selling 20 candles: 20 candles * \$2.00/candle = \$40.00. \\ 
  Now, let's calculate the total profit: Total revenue - Total cost = \$40.00 - \$10.00 = \$30.00. \\
  Answer: \$30.00 \textcolor{red}{\XSolidBrush}
  \tcblower
  \textbf{P3 system prompt}: You are a mathematics expert tasked with solving a variety of math problems. \textcolor{purple}{Upon receiving a user query, carefully analyze the provided hints in the complementary instruction to guide your solution process}. Offer a detailed, step-by-step explanation of your approach, ensuring clarity and coherence throughout. Always conclude the last line of your response should be of the following format: "Answer: \$VALUE" where VALUE is a numerical value. \\
  \textbf{P3 user prompt}: \#\#\# User prompt: \\
  Charlie wants to sell beeswax candles. For every pound of beeswax, he can make 10 tapered candles. One pound of beeswax and the wicks cost \$10.00 in supplies. If he sells each candle for \$2.00 each, what is his net profit if he makes and sells 20 candles? \\
  \#\#\# Complementary instruction: \\
   \colorbox{pink}{First, calculate the total cost of materials for making 20 candles},\colorbox{yellow}{then determine the} \colorbox{yellow}{total revenue from selling those candles}.\colorbox{lime}{Finally, subtract the total cost from the total} \colorbox{lime}{revenue to find the net profit}.\\
  \textbf{P3 answer}: To find Charlie's net profit, we need to calculate the total cost of materials for making 20 candles, then determine the total revenue from selling those candles, and finally subtract the total cost from the total revenue.\\
  \colorbox{pink}{First, let's calculate the total cost of materials for making 20 candles}:\\
  - 1 pound of beeswax and wicks cost \$10.00 \\
  - For 20 candles, he needs 2 pounds of beeswax (20 candles / 10 candles per pound) \\
  - The cost of 2 pounds of beeswax and wicks is 2 * \$10.00 = \$20.00 \\
  \colorbox{yellow}{Next, let's determine the total revenue from selling 20 candles}: \\
  - He sells each candle for \$2.00 \\
  - Total revenue = 20 candles * \$2.00/candle = \$40.00 \\
  \colorbox{lime}{Now, let's calculate the net profit}: \\
  - Net profit = Total revenue - Total cost \\
  - Net profit = \$40.00 - \$20.00 = \$20.00 \\  
  So, Charlie's net profit from making and selling 20 candles is \$20.00.
  
  Answer: \$20.00 \textcolor{red}{\CheckmarkBold}
\end{tcolorbox}
}
\caption{Comparison between PAS prompts and P3 prompts.}
\label{fig:case_affinity}
\end{figure}

\clearpage

\subsection{Algorithm Pseudocode}
\label{appendix:algo}

\begin{algorithm}[htb]
    \caption{P3 Offline Optimization}
    \label{alg1}
    \begin{algorithmic}
    \REQUIRE Query dataset $X$, current system prompt $x_s$, search depth D, system prompt optimization interval T, good query buffer $X_\mu$, hard query buffer $X'_\mu$
    \FOR{ $x$ in $X$}
        \STATE Generate $k$ candidate complements $\{e_j\}_{j=1:k}$
        \STATE Generate answers for those candidate complements with $x$ and $x_s$ 
        \STATE Evaluate answers using LLM-as-judge, obtain scores $\{s_j\}_{j=1:k}$ 
        \FOR{d = 1 to D}
            \STATE Use $\{(e_j, s_j)\}_{j=1:k}$ as examplers to few-shot generate new candidates $\{e_{k+j}\}_{j=1:c}$
            \STATE Generate answers for new candidate complements with $x$ and $x_s$
            \STATE Evaluate new answers using LLM-as-judge, obtain scores $\{s_{k+j}\}_{j=1:c}$ 
            \STATE Select the top $k$ candidates in $\{e_j\}_{j=1:k+c}$
        \ENDFOR
        \STATE Choose the best of those candidates as $e^{*}$, with its corresponding score $s^{*}$
        \IF{ $s^{*}$ > threshold $\epsilon$}
            \STATE Append ($x$, $e^{*}$) to $X'_\mu$
        \ELSE
            \STATE Append ($x$, $e^{*}$) to $X_\mu$
        \ENDIF
        \IF {index(x) \% T == 0}
            \STATE Optimize system prompt with Algorithm~\ref{alg:sys} 
        \ENDIF
    \ENDFOR
    \end{algorithmic}
\end{algorithm}
\vspace{0em}
\begin{algorithm}[H]
\caption{P3 System Prompt Optimization}
\label{alg:sys}
    \begin{algorithmic}[1]
    \REQUIRE current system prompt $x_s$, hard query buffer $X_\mu$, system prompt buffer $X_s$
        \STATE Sample a batch $B$ from $X_\mu$
        \IF{ $X_s$ is empty}
            \STATE Refine $x_s$ to generate k candidate system prompts
            \STATE Evaluate those candidates on $B$
            \STATE Update $x_s$ to the best candidate
            \STATE Add those candidates to $X_s$
        \ELSE
            \STATE Sample $\{x_s^j\}_{j=1:k}$ from $X_s$ with replacement
            \STATE Evaluate those samples on $B$, obtain scores $\{s_j\}_{j=1:k}$
            \STATE Use $\{(x_s^j, s_j)\}_{j=1:k}$ as examplers to few-shot generate $c$ new system prompts $\{x_s^{k+j}\}_{j=1:c}$
            \STATE Evaluate new system prompts on $B$, obtain scores $\{s_{k+j}\}_{j=1:c}$
            \STATE Update $x_s$ to the best of $\{x_s^j\}_{j=1:k+c}$
            \STATE Select top $C$ new system prompts and append to $X_s$
        \ENDIF
    \end{algorithmic}
\end{algorithm}

\subsection{Hyperparameters for Offline Optimization}
\label{appendix:hyper}
We set $\epsilon=6$, $k=5$, $c=5$ and $C=3$ for all the experiments. We set $T=400$ for PAS and GSM8k datasets, and $T=80$ for the GPQA dataset. Here $\epsilon$ is an empirical number which we adjust based on the difficulty of the dataset.

\clearpage

\subsection{Meta Prompts for Offline Optimization}
\label{appendix:meta}
\begin{tcolorbox}[colback=blue!5!white,colframe=blue!75!black,title=Meta Prompt for Complement Generation]
\textbf{System}: You are a proficient prompt engineer. Your task is to add a complementary instruction at the end of the user's prompt to enhance the quality of responses from a large language model. \\

\#\#\# Note: \\
1. The instruction should serve as a guidance for the answer, providing feasible direction and thought process, but not revealing the answer itself. \\
2. The complementary instruction should focus on general methodology. Try to keep it short and concise. \\
3. The complementary instruction should be in the same language with the user prompt. \\
4. When generating your response, the complementary instruction should be bracketed with <INS> and </INS>.
\tcblower
\textbf{User}: <user\_prompt>{prompt}</user\_prompt>
\end{tcolorbox}

\begin{tcolorbox}[colback=blue!5!white,colframe=blue!75!black,title=Meta Prompt for Complement Optimization]
Your task is to generate an instruction as a complementary text for the user prompt. \\

\#\#\# User prompt: \\
\{prompt\} \\

Below are some previous instructions with their scores. The score ranges from 0 to 10. Higher score indicates higher quality. \\

\#\#\# Examples: \\
\{examplers\} \\

Now generate a new complementary instruction that is superior than all the instructions above. The instruction should serve as a guidance for the answer, providing feasible direction and thought process, but not revealing the answer itself. Try to keep it short and concise. The complementary instruction should begin with <INS> and end with </INS>.
\end{tcolorbox}

\begin{tcolorbox}[colback=blue!5!white,colframe=blue!75!black,title=Meta Prompt for LLM-as-judge]
\textbf{System}: Please act as an impartial judge and evaluate the quality of the response provided by an AI assistant to the user question displayed below. Your evaluation should consider factors such as the helpfulness, relevance, accuracy, depth, creativity, and level of detail of the response. Begin your evaluation by providing a short explanation. Be as objective as possible. After providing your explanation, please rate the response on a scale of 1 to 10. The rating should begin with <score> and end with </score>, for example: "<score>5</score>".
\tcblower
\textbf{User}: <Question>:\\
\{question\}\\\\
<Assistant's Answer>:\\
\{answer\}
\end{tcolorbox}

\vspace{-5em}
\begin{tcolorbox}[colback=blue!5!white,colframe=blue!75!black,title=Meta Prompt for LLM-as-judge with Reference Answer]
\textbf{System}: Please act as an impartial judge and evaluate the quality of the response provided by an AI assistant to the user question displayed below. Your evaluation should focus on correctness. You will be given the original question, the assistant's answer and the reference answer. Begin your evaluation by comparing assistant's final answer with the reference answer. Then check the correctness of the reasoning steps in assistant's answer. After providing your explanation, please rate the response on a scale of 1 to 10. Correct answers should receive high scores while false answers should receive low scores. The rating should begin with <score> and end with </score>, for example: "<score>5</score>".
\tcblower
\textbf{User}: <Question>:\{question\}\\\\
<Assistant's Answer>:\{answer\}\\\\
<Reference Answer>:\{reference\}
\end{tcolorbox}

\begin{tcolorbox}[colback=blue!5!white,colframe=blue!75!black,title=Meta Prompt for System Prompt Generation]
You are a large language model prompt engineer, responsible for completing prompt optimization tasks. Rewrite and refine user input prompts based on criteria such as usefulness (e.g., depth, creativity, coherence), relevance, and fluency. \\

\#\#\# User Prompt: \\
\{prompt\} \\

\#\#\# Important Notes: \\
1. The optimized prompt should help LLMs to provide better answers for a wide range of queries. \\
2. There is always a piece of complementary instruction in user queries, which provides hints to solve the queries. Make sure to make good use of this information. \\
3. When generating your response, the optimized prompt should be bracketed with <INS> and </INS>.
\end{tcolorbox}

\begin{tcolorbox}[breakable,colback=blue!5!white,colframe=blue!75!black,title=Meta Prompt for System Prompt Optimization]
You are a large language model prompt engineer responsible for completing prompt optimization tasks by using historical prompt optimization records to derive more effective prompts. \\

Below are some historically optimized prompts along with their scores. The score range is from 0 to 10, with higher scores indicating better quality. \\\\
\{examplers\} \\

\#\#\# Important Notes: \\
1. Analyze the evolution patterns and score trends of historical prompts, and summarize the effective optimization rules. \\
2. Based on the summarized optimization rules, generate a new prompt that is superior than all the previous ones. \\
3. There is always a piece of complementary instruction in user queries, which provide hints to solve the queries. Make sure to make good use of this information. \\
4. The newly generated prompt must be bracketed with <INS> and </INS> tags.
\end{tcolorbox}

\subsection{Details of Training Online Query Optimization Model}
\label{appendix:online}
As for training the online query optimization model, we finetune a Qwen2-7b-instruct model on the offline collected (user prompt, complement) pairs. 

For the PAS dataset, we employ full-parameter finetuning with DeepSpeed Zero~\cite{rajbhandari2020zero}. The optimizer is AdamW, and the learning rate is set to 1e-5. We collect 7605 samples out of 7729 in the original dataset, the rest is filtered as hard samples for system prompt optimization.

For the GSM8k dataset, we also employ full-parameter finetuning with DeepSpeed Zero~\cite{rajbhandari2020zero}. The optimizer is AdamW, and the learning rate is set to 1e-5. We collect 7361 samples out of 7473 in the original training dataset (\textsf{train.json}), the rest is filtered as hard samples for system prompt optimization.

For the GPQA dataset, since the volume of this dataset is much smaller, we employ LoRA plus~\cite{hayou2024loraplus} as the SFT method, with rank r = 8, learning rate lr = 3e-4 and LoRA+ ratio $\lambda=16$. The optimizer is also AdamW. We create a training dataset by fetching unduplicated samples from \textsf{gpqa\_main.csv} and \textsf{gpqa\_extended.csv}, then removing the duplicated samples in \textsf{gpqa\_diamond.csv}. We collect 299 samples out of 348 in this training set. And the rest is filtered as hard samples for system prompt optimization.

For both training and inference, we set a system prompt for the online optimization model as follows:

\begin{tcolorbox}[breakable,colback=blue!5!white,colframe=blue!75!black,title=System Prompt for Online Optimization Model]
You are an expert in enhancing user prompts, and your task is to add a supplementary prompt at the end of the user's prompt to enhance the quality of responses from a large language model. \\

Note: The supplementary prompt should serve as a guide for the response, providing accurate direction and thought process for the actual answer, but must not give away the answer itself. The supplementary prompt should be concise and effective.
\end{tcolorbox}

\end{document}